\documentclass[letterpaper]{article} 
\usepackage{aaai2026}  
\usepackage{times}  
\usepackage{helvet}  
\usepackage{courier}  
\usepackage[hyphens]{url}  
\usepackage{graphicx} 
\usepackage{natbib}  
\usepackage{caption} 
\usepackage{algorithm}
\usepackage{algorithmic}
\usepackage{tikz}
\usepackage{booktabs}
\usepackage{tikz}
\usetikzlibrary{mindmap,trees}
\usepackage{enumitem}
\setlist[itemize]{leftmargin=*}

\usepackage{newfloat}
\usepackage{listings}
\usepackage{verbatim}
\DeclareCaptionStyle{ruled}{labelfont=normalfont,labelsep=colon,strut=off} 
\floatstyle{ruled}
\newfloat{listing}{tb}{lst}{}
\floatname{listing}{Listing}
\title{From Benchmarks to Business Impact: Deploying IBM Generalist Agent in Enterprise Production}
\author {
    Segev Shlomov\textsuperscript{\rm 1},
    Alon Oved\textsuperscript{\rm 1},
    Sami Marreed\textsuperscript{\rm 1},
    Ido Levy\textsuperscript{\rm 1},
    Offer Akrabi\textsuperscript{\rm 1},
    Avi Yaeli\textsuperscript{\rm 1},
    \L ukasz Strak\textsuperscript{\rm 2},
    Elizabeth Koumpan\textsuperscript{\rm 2},
    Yinon Goldshtein\textsuperscript{\rm 1},
    Eilam Shapira\textsuperscript{\rm 1},
    Nir Mashkif\textsuperscript{\rm 1},
    Asaf Adi\textsuperscript{\rm 1}
}
\affiliations {
    \textsuperscript{\rm 1}IBM Research \quad
    \textsuperscript{\rm 2}IBM Consulting \\
    \texttt{\{segev.shlomov1, alon.oved, sami.marreed, ido.levy1\}@ibm.com, \\ 
    \{offer.akrabi, aviy\}@il.ibm.com, \\
    \{lukasz.strak@pl.ibm.com, ekoumpan@ca.ibm.com\}, \\
    \{yinong, eilam.shapira\}@ibm.com, \{nirm, adi\}@il.ibm.com}
    
}

\begin{document}

\maketitle

\begin{abstract}
Agents are rapidly advancing in automating digital work, but enterprises face a harder challenge: moving beyond prototypes to deployed systems that deliver measurable business value. This path is complicated by fragmented frameworks, slow development, and the absence of standardized evaluation practices. Generalist agents have emerged as a promising direction, excelling on academic benchmarks and offering flexibility across task types, applications, and modalities. Yet, evidence of their use in production enterprise settings remains limited. This paper reports IBM’s experience developing and piloting the Computer Using Generalist Agent (CUGA), which has been open-sourced\footnote{\url{https://github.com/cuga-project/cuga-agent}} for the community. CUGA adopts a hierarchical planner--executor architecture with strong analytical foundations, achieving state-of-the-art performance on AppWorld and WebArena. Beyond benchmarks, it was evaluated in a pilot within the Business-Process-Outsourcing talent acquisition domain, addressing enterprise requirements for scalability, auditability, safety, and governance. To support assessment, we introduce \textbf{BPO-TA}, a 26-task benchmark spanning 13 analytics endpoints. In preliminary evaluations, CUGA approached the accuracy of specialized agents while indicating potential for reducing development time and cost. Our contribution is twofold: presenting early evidence of generalist agents operating at enterprise scale, and distilling technical and organizational lessons from this initial pilot. We outline requirements and next steps for advancing research-grade architectures like CUGA into robust, enterprise-ready systems.
\end{abstract}

\section{Introduction}

Enterprises are under growing pressure to automate digital work at scale. From customer support to back-office analytics, knowledge workers routinely interact with heterogeneous environments--web portals, APIs, spreadsheets, and dashboards--while facing strict requirements for auditability, reproducibility, privacy, and cost control. Over the past two years, interest has surged in \emph{computer-using agents} (CUAs), systems that can plan and execute multi-step tasks across diverse applications. Yet for enterprises, the challenge is not only to prove capability--it is to \emph{productize agents and capture real business value}.

\textbf{The enterprise need.}
In practice, organizations struggle with the journey from research to deployment. Multiple frameworks and architectural patterns compete for adoption, but few offer clear guidance on speed of development, time-to-value, cost efficiency, and reliability in production settings. Enterprises also lack standardized ways to evaluate agentic systems: benchmarks emphasize academic settings, while business leaders demand measurable impact such as SLA compliance, reduction in manual effort, or improved audit readiness. Bridging this gap requires not only technical advances in agent design but also organizational insights into deployment, governance, and monitoring.

\textbf{The research gap.}
Recent work has shown that \emph{generalist} agents--single systems designed to perform diverse computer-use tasks--can achieve impressive results on academic benchmarks. Generalist designs are attractive because they promise (i) adaptability across task types and domains, (ii) reusability of architecture and tooling across new environments, and (iii) reduced need for brittle, task-specific scripting. However, published research has so far remained benchmark-centric. Generalist agents have been shown to perform well on synthetic benchmarks, yet their effectiveness in \emph{enterprise production settings} is largely untested. The central question is therefore: what modifications, safeguards, and evaluation methods are required to make generalist agents \emph{enterprise-ready}?

\textbf{Our approach.}
This paper addresses that question through IBM’s development of CUGA. Architecturally, CUGA evolved into a hierarchical planner–executor system with three control layers: (i) a chat/context layer for preprocessing inputs, (ii) an outer loop for task planning and management using a persistent ledger, and (iii) an inner loop for sub-task execution via specialized agents (API, Web, CLI, file-system). Reliability mechanisms include schema-grounded prompting, variable tracking, reflective retries, and provenance logging. In its benchmark evaluation, CUGA achieved state-of-the-art performance on both AppWorld and WebArena, confirming the strength of its generalist design.
But more importantly for enterprises, CUGA was piloted in the Business-Process-Outsourcing (BPO) talent acquisition domain--a setting where recruiters and analysts must answer evidence-based questions across multiple dashboards and datasets under policy and audit constraints. We use this pre-deployment pilot not as the main “story,” but as a proving ground to evaluate what enterprise readiness demands.

We introduce a new domain-specific benchmark, \textbf{BPO-TA}, comprising 26 decision-support tasks across 13 read-only analytics endpoints. Tasks span single-endpoint lookups, cross-API joins, provenance-grounded explanations, and the graceful handling of unsupported queries. This benchmark enabled both regression testing and controlled ablation studies during CUGA’s pilot evaluation. In preliminary tests within simulated enterprise workflows, CUGA approached the accuracy of hand-crafted agents while indicating potential substantial reductions in development time (up to 90\%) and cost (up to 50\%). These early findings suggest that generalist designs can enable measurable enterprise value when adapted with appropriate safeguards. 

This paper contributes:
\begin{itemize}
\item \textbf{Enterprise pilot experience.} Evidence from a pilot of a generalist agent evaluated with recruiters and analysts in the BPO talent acquisition domain, including architectural modifications for auditability, safety, and governance.
\item \textbf{Domain benchmark.} The \textbf{BPO-TA} benchmark that captures realistic enterprise analytics queries, enabling reproducible regression testing and ablation studies.
\item \textbf{Architectural advances.} A planner–executor agent design with schema-grounded prompting, variable tracking, reflective retries, provenance logging, and an API/Tool Hub that streamlined onboarding of enterprise applications. This architecture achieved state-of-the-art performance on both WebArena and AppWorld benchmarks.
\item \textbf{Preliminary business impact.} Early evaluations showed accuracy approaching that of hand-crafted agents, with indications of up to 90\% reduction in development time and 50\% reduction in development cost, alongside improved time-to-answer and reproducibility.
\item \textbf{Lessons learned.} Technical and organizational insights from the pilot, including monitoring, governance alignment, and maintenance practices required to transition generalist agents from research to enterprise readiness.
\end{itemize}

\section{Related Work}

Early agentic paradigms such as \textit{ReAct} interleave chain-of-thought reasoning with environment actions to improve task completion and interpretability \cite{yao2023react}, while code-centric approaches as \textit{CodeAct} generate executable code to plan and call tools/APIs for complex tasks \cite{ye2024codeact}. These ideas catalyzed practical enterprise frameworks that orchestrate multiple specialized agents (or tools) with configurable roles. From AutoGen's conversation-programmed multi-agent patterns \cite{wu2023autogen}, to LangGraph's stateful, tool-grounded agent graphs for reliability and recoverability \cite{langgraph2024}, and OpenAI's \emph{Swarm} orchestrator for multi-agent handoffs \cite{openaiSwarm2024}. Despite promise, production experience consistently reports fragility at scale: brittle inter-agent handoffs, maintenance overhead from prompt/tool drift, safety and generalization across different domains.

Enterprise report measurable wins when agentic systems are embedded behind assist and self-serve flows \cite{verizonVerge2025,verizonMWL2025, aibusinessAlibaba2022}. Analytics/BI copilots ship text-to-SQL/report-generation agents tightly coupled to enterprise data governance, observability, and review workflows \cite{snowflakeCortex2024,msPowerBI2024,databricksGenie2025}. Broader surveys in hiring/HR analytics highlight fairness, transparency, and audit requirements in employment contexts, motivating provenance and explainability \citep{raghavan2020mitigating,fabris2025fairness,chen2023ethics,schwartz2023enhancing}. Industry adoption reports likewise stress governance, measurable ROI, and operational readiness (monitoring, latency/cost budgets) as prerequisites for scale-out \citep{zhang2025agent,cemri2025multi}. These constraints shape the design space of enterprise agents task completion. BPO process automation combine workflow orchestration, retrieval over enterprise knowledge, and document understanding to automate outsource non-core business functions to third-party providers, leveraging semantic reasoning for process understanding as explored in \citep{oved2025snap}, and human-centric automation approaches such as IDA and conversational RPA \citep{shlomov2024ida,yaeli2022recommending,zeltyn2022prescriptive}.

Concurrently, the research community has pushed toward generalist CUAs that plan and act across heterogeneous software. WebArena offers realistic, self-hosted websites for browser agents and showed early baselines struggled to exceed modest end-to-end success \cite{zhou2023webarena}. AppWorld evaluates multi-application orchestration via hundreds of programmatic APIs with outcome-based grading \citep{trivedi2024appworld}. OSWorld measures GUI workflows on desktop applications and OS tasks \citep{xie2024osworld}. Complementary suites probe interaction quality and oversight: \textit{ST-WebAgentBench} emphasizes policy adherence in web agents, introducing \emph{Completion-under-Policy} (CuP) as the primary objective~\cite{levy2025stwebagentbench}, 
$\tau$-Bench targets tool-agent-user dynamics and policy/instruction following \citep{yao2024tau},
BrowserGym provides a unified platform for evaluating web agents under controlled variability \citep{debrowsergym} while \citep{shlomov2024grounding} identifies planning as the dominant bottleneck in web agents. Architecturally, generalist CUAs increasingly combine hierarchical planning, explicit state/variable tracking, and reflective repair during execution to improve robustness in long-horizon settings \citep{shinn2023reflexion, kim2023language, zhang2025agentorchestra}. Recent vendor-facing systems expose ``computer use'' capabilities (desktop/browser control, file ops) under sandboxes, signaling a trend toward production CUAs \citep{anthropic2024computeruse,openai2025operator, shen2025mind, fourney2024magentic}.

Despite rapid progress, several gaps limit direct transfer from benchmark success to enterprise deployment. First, \emph{governance}: high-risk domains (including employment analytics) demand provenance, HITL, and post-deployment monitoring mandated by frameworks and regulation \citep{nist2023airmf,euai2024,nyc144}. Second, \emph{tool proliferation and schema variance}: results degrade as agents shortlist from dozens of APIs and maintain consistency across dependent calls \citep{shen2024taskbench,xu2023tool}. Third, \emph{operational constraints}: latency, cost, and reproducibility must be tracked and controlled in production \citep{kwon2023efficient,zheng2024sglang,jiang2023llmlingua}. Fourth, \emph{graceful degradation}: enterprise agents must decline unsupported requests without hallucination, and surface transparent rationales and computation logs \citep{nakano2021webgpt}. Our work addresses this bridge by showing how a generalist CUA can be adapted for enterprise: schema-minimized API onboarding, deterministic parsing/validation, provenance-first responses, in a piloted BPO talent acquisition setting.


\begin{figure*}[ht!]
    \centering
    \includegraphics[width=0.9\linewidth]{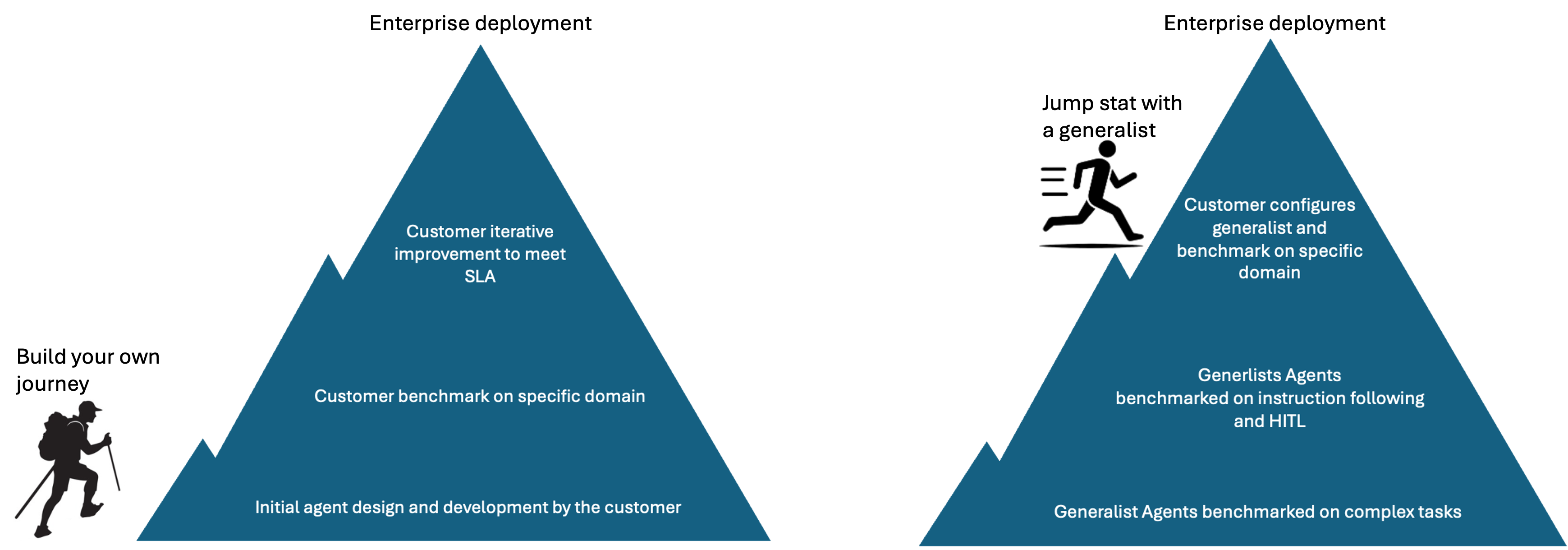}
    \caption{Reducing time-to-value with generalist agents. Traditional specialized agents (left) require extensive custom design and benchmarking. Generalist agents (right), benchmarked on complex tasks such as AppWorld, WebArena, and TauBench2, shift the enterprise focus to configuration and domain-specific evaluation.}
    \label{fig:time-to-value-generalist}
\end{figure*}
\section{The Application: BPO Talent Agent}

\textbf{Background and Business Context}
IBM Consulting operates a double-digit million business in Business Process Outsourcing (BPO) for Talent Acquisition (TA). In this model, IBM specialists manage recruitment pipelines on behalf of client organizations, often working across multiple HR platforms, analytics dashboards, and reporting tools. While effective, the manual workflow is labor-intensive: recruiters spend significant time pulling data, reconciling spreadsheets, and preparing insights for hiring managers. Service-level agreements (SLAs) around time-to-hire, conversion funnels, and sourcing performance are business-critical, yet measuring and optimizing them has historically been slow and error-prone.  

The vision behind the Talent Acquisition Agent was to augment human recruiters rather than replace them--acting as a digital sidekick that can provide proactive insights, automate repetitive analysis, and surface SLA risks before they become client issues. A core design principle is \textbf{human-in-the-loop (HITL)}: the business configures where the agent can act autonomously and where human oversight is mandatory, and the agent must strictly adhere to these requirements.

\subsection{The Development Journey of Agentic Systems}
The BPO--TA project also reflects a broader pattern observed across enterprises experimenting with agentic systems. Most teams begin with quick wins: popular frameworks like \textbf{ReAct} or \textbf{CodeAct} can be instantiated in days, yielding impressive demos where agents call APIs or generate code to answer queries. But as projects scale, limitations surface:
\begin{itemize}
    \item ReAct agents degrade when required to juggle more than a handful of tools.
    \item Developers patch around this by introducing routers and delegators, creating ``white-box'' architectures with fragile hand-offs between sub-agents.
    \item Complex instructions or policy requirements (e.g., privacy, governance) strain these prototypes further.
    \item Roadmaps become unclear, with teams caught in cycles of experimentation rather than predictable progress.
\end{itemize}

The BPO--TA team followed this arc. An initial prototype, built quickly on reactive patterns, showed promise but could not scale to the breadth of sources and policies in Talent Acquisition. With 13 APIs spanning multiple systems and providers--each offering several actions and requiring orchestration across workflows--the complexity exceeded what the prototype could handle. At this inflection point, the team turned to IBM Research to evaluate CUGA, which had just achieved state-of-the-art results on AppWorld and WebArena benchmarks. The open question was: \emph{Could a benchmark-proven generalist agent deliver enterprise-grade performance in the demanding Talent Acquisition setting?}

\subsection{Application Setup}
The Talent Acquisition Agent is still on its journey toward full production deployment. To date, it has achieved on-par accuracy with specialized agents, is being evaluated against enterprise requirements, and is under consideration for production rollout. The deployment context includes several key characteristics:
\begin{itemize}
    \item \textbf{APIs and Analytics Layer:} The environment exposes 13 read-only APIs, spanning multiple applications and providers. Each API offers several actions, and realistic workflows often require orchestrating across sources. Examples include SLA-by-source, funnel conversion, hires by percentage, and skill-impact analysis.
    \item \textbf{Governance and Security:} To build trust at low risk, the current configuration is restricted to read-only APIs. This allows experimentation and validation without impacting underlying systems. Over time, the goal is to progress toward create/update capabilities, enabling fully automated workflows once safety and trust are established. All responses include provenance logs, and PII is excluded or redacted to maintain compliance.
    \item \textbf{Integration with Business Workflows:} The agent is designed to embed into recruiters' existing dashboards in the browser, becoming part of their daily workflow rather than a separate tool. It must integrate seamlessly with the user experience and identity controls already in place.
    \item \textbf{Human + Agent Collaboration:} Recruiters and analysts interact with the agent through a conversational interface. Depending on business configuration, the agent may act autonomously or defer decisions back to humans. HITL requirements are explicit and configurable, ensuring alignment with business workflows and governance.
\end{itemize}

\subsection{Why BPO--TA Matters as a Testbed}
The BPO--TA pilot illustrates why Talent Acquisition is a representative domain for studying the enterprise readiness of generalist agents:
\begin{itemize}
    \item \textbf{Complex orchestration:} Multi-source workflows spanning 13 APIs, each with multiple actions, often requiring reasoning across providers and data sources.
    \item \textbf{Governance-heavy:} Read-only experimentation mode, HITL oversight, audit trails, and compliance constraints.
    \item \textbf{High value:} A double-digit million business unit where even modest efficiency gains translate to significant client impact.
    \item \textbf{Scalable lesson:} The trajectory from quick prototypes to generalist adoption mirrors what many enterprises experience in their agent journey.
\end{itemize}

In short, BPO--TA provided the ideal proving ground: a live enterprise context, business-critical stakes, and a context where the shortcomings of early architectures were well understood. The Talent Acquisition Agent was not introduced into a greenfield environment--it was evaluated precisely at the point where conventional approaches had reached their limits. This made the journey both realistic and consequential: success here could validate generalist architectures and signal how such systems may bridge the gap between academic benchmarks and enterprise deployment.

\section{Enterprise Requirements for Generalist Agents}

From the BPO--TA use case described earlier, and from additional discussions with other business units in our organization--including Finance, Sales, Procurement, Legal, and the CIO’s office--we observe a recurring pattern. Enterprises consistently identify a set of requirements that go beyond academic benchmarks. In this section, we summarize our understanding of what is needed from generalist agents in the enterprise.

\paragraph{Safety and Trustworthiness (Top Priority).}  
Enterprises adopt agents in production only when their safety and trustworthiness are ensured, including:
\begin{itemize}
    \item \textbf{Instruction following and policy adherence:} agents must reliably comply with organizational rules, workflows, and domain-specific processes.
    \item \textbf{Transparency and consistency:} outputs should be reproducible, grounded in provenance, and free of unexplained variability.
    \item \textbf{Avoidance of hallucinations:} agents must not fabricate data or invent actions in business-critical workflows.
    \item \textbf{Configurable human-in-the-loop (HITL):} oversight must be adjustable by the business--defining where autonomy is permitted and where human approval is mandatory.
    \item \textbf{Security baseline:} restricted access, provenance logging, and minimal permissions sufficient to comply with enterprise governance frameworks.
\end{itemize}

\paragraph{Efficiency and Cost-Performance.}  
Once accuracy and safety are established, efficiency becomes the next priority. Agents must deliver results with acceptable latency and without prohibitive compute costs. Token efficiency, reduced retries, and optimized execution are crucial to ensuring that deployments scale economically.

\paragraph{Integration and Context-Awareness.}  
Agents should integrate directly into existing business workflows and user experiences, such as dashboards or browser-based recruiter tools, rather than creating separate silos. They must also be context-aware, recognizing what the user is seeing or doing, in order to provide relevant support without forcing disruptive context switching.

\paragraph{Policy Alignment and Instruction Following.}
Enterprise processes are rarely generic. In Talent Acquisition, for example, requisition workflows and SLA definitions act as concrete guardrails that agents must follow. Effective alignment requires agents to learn from enterprise documentation (such as playbooks, policies, and guidelines written in natural language), from user demonstrations that model correct behavior, and from ongoing feedback that enables them to gradually refine their responses and conform to organizational norms.

\subsection{The Value of Generalist Agents}

Generalist agents provide a promising foundation for these requirements. Unlike specialized agents that must be hand-crafted for each domain, generalists are trained on diverse benchmarks covering complex task completion, reasoning, and instruction following. For example, benchmarks such as AppWorld, WebArena, and TauBench2 evaluate tool use, multi-turn reasoning, instruction following, and even human-in-the-loop interaction.

This foundation allows enterprises to focus not on building agents from scratch, but on \textbf{configuring and benchmarking} the agent for their specific domain. Instead of months of custom design and iterative experimentation, organizations can move to value within weeks. Generalist agents reduce:
\begin{itemize}
    \item \textbf{Time-to-value:} shifting from a 3--9 month development cycle to a few weeks of configuration and testing.
    \item \textbf{Development effort:} enabling enterprises to inherit baseline capabilities in accuracy, instruction following, and safety.
    \item \textbf{Risk:} lowering the likelihood of project fatigue and failed adoption, since much of the heavy lifting has already been validated on foundation benchmarks.
\end{itemize}

Figure~\ref{fig:time-to-value-generalist} illustrates this contrast: while traditional specialized agents require extensive design, custom benchmarks, and iterative refinement before deployment, generalist agents like CUGA allow enterprises to inherit strong foundations and reach deployment readiness with far less effort.

\section{System Architecture and Pre-deployment}
\label{sec:system}

\begin{figure}[t]
  \centering
  \includegraphics[width=\linewidth]{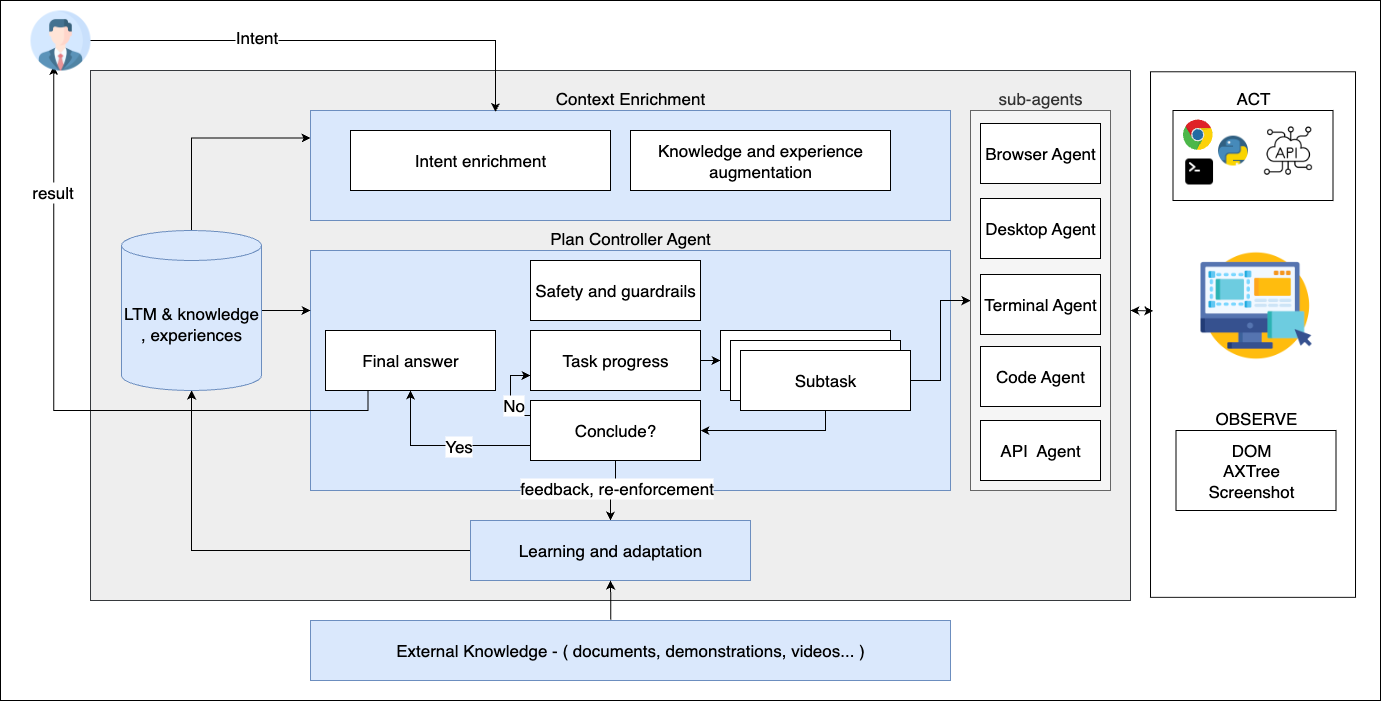}
  \caption{CUGA general architecture via Layered planner–executor loops.}
  \label{fig:layers}
\end{figure}

\paragraph{Layered planner–executor loops.}
CUGA implements a hierarchical agentic architecture \cite{marreed2025towards} with nested planner–executor loops (Fig.~\ref{fig:layers}). At the top, an optional \emph{chat layer} provides input interpretation and lightweight preprocessing, including message and variable histories; this can be bypassed in non-chat deployments. The \emph{outer loop} governs \emph{task planning and orchestration}: a \emph{Task Analyzer} identifies the target application, a \emph{Task Decomposer} determines whether multi-application coordination is required, and a persistent \emph{Plan Controller} advances a durable \emph{task ledger}. The ledger records steps, variable bindings, replans, and completions. In pilot evaluations, this ledger was essential for traceability, compliance, and recovery from partial failures. The \emph{inner loop} delegates sub-tasks to specialized agents--\emph{API/Tool}, \emph{Web Browser}, \emph{CLI}, and domain-specific agents--each acting within its own environment and returning structured observations to the controller.

\paragraph{Planner-centric sub-agents.}
Two execution families illustrate the planner–executor pattern: the \emph{API Sub Agent} and the \emph{Browser Sub Agent}.
The API Sub Agent combines short-term memory, an API Planner, and strategic reflection, coordinating a \emph{ShortlisterAgent} (which selects APIs through a registry) and a \emph{CodeAgent} (with a nested \emph{CodePlanner} and sandboxed executor). This modular design allowed rapid onboarding of analytics endpoints in the BPO pilot.  
The Browser Sub Agent pairs a \emph{Browser Planner} and Reflection Judge with two execution paths: an \emph{Action Agent} (click, type, select, navigate) and a \emph{Question Answering Agent} (DOM-to-Markdown conversion and screenshots). Although disabled in the BPO deployment for governance reasons, this design enables seamless switching between API-first and hybrid browser–API workflows without re-architecting.

\paragraph{Reliability: task ledger, interrupt nodes, reflective retries.}
The Plan Controller enforces reliability by schema-grounded prompting, validation, and explicit \emph{Interrupt Nodes}. When tool responses deviate from schema or produce unexpected results, reflective checks are invoked and invalid plans or parameters are repaired before resuming execution. This cycle of prompt $\rightarrow$ call $\rightarrow$ validation $\rightarrow$ reflection/replan reduced parsing-related failures by more than one-third in internal pilot runs.

\paragraph{API/Tool Hub and schema standardization.}
To scale beyond prototypes, CUGA replaced per-application MCP servers with a centralized \emph{API/Tool Hub}. The hub minimizes OpenAPI specifications into LLM-friendly schemas, canonicalizes parameter names and types, attaches domain-specific notes, and enforces strict JSON-schema I/O. This eliminated per-app server maintenance and reduced onboarding time for new endpoints from weeks to hours.

\paragraph{Sandboxed computation for safety.}
For lightweight computation (joins, aggregations, deltas), the \emph{API--Code path} generates structured pseudo-code via a \emph{CodePlanner}, executed inside a restricted \emph{Code Agent sandbox}. The sandbox isolates file/network access, enforces execution budgets, and logs all computations for audit. This allowed domain analysts to use AI-augmented workflows without compromising governance or data-handling policies.

\paragraph{Web agent path and governance.}
The architecture also supports a \emph{Web Planner Agent} coordinating \emph{Web Action} and \emph{Web Q\&A} sub-agents in a Playwright–Chromium runtime. In the BPO pilot, this capability was deliberately disabled to comply with enterprise governance requirements, but the design demonstrates flexibility for hybrid deployments that combine API-first and web-based workflows.

\section{Benchmarks and Pilot Evaluation}
\label{sec:evaluation}

\paragraph{State-of-the-art Benchmarks.}
CUGA achieves state-of-the-art results on both WebArena and AppWorld, ranking first among published agents. Tables~\ref{tab:webarena} and \ref{tab:appworld} present detailed per-application and per-level outcomes. On WebArena, CUGA attains an overall accuracy of \textbf{61.7\%}, with the strongest performance on Reddit (75.5\%) and Map (64.2\%). On AppWorld's "Test-Challenge" dataset, CUGA reaches \textbf{48.2\%} overall scenario completion, with particularly high success on Level~1 tasks (87.5\% scenario completion). These results demonstrate that the CUGA architecture is competitive with--and in many cases surpasses--specialized systems. More details can be found in the Appendix.

\begin{table}[ht!]
    \centering
    \caption{Performance of CUGA on the \textbf{WebArena} benchmark.}
    \begin{tabular}{l c}
        \toprule
        \textbf{Application} & \textbf{Accuracy (\%)} \\
        \midrule
        GitLab          & 61.7 \\
        Map             & 64.2 \\
        Reddit          & 75.5 \\
        Shopping        & 58.3 \\
        Shopping Admin  & 62.6 \\
        Multi-App       & 35.4 \\
        \midrule
        Overall         & \textbf{61.7} \\
        \bottomrule
    \end{tabular}
    \label{tab:webarena}
\end{table}


\begin{table}[ht!]
    \centering
    \caption{Performance of CUGA on \textbf{AppWorld}.}
    \resizebox{\linewidth}{!}{%
    \begin{tabular}{l cc cc cc}
        \toprule
        {\textbf{Level}} &
        \multicolumn{2}{c}{\textbf{Task Goal (\%)}} &
        \multicolumn{2}{c}{\textbf{Scenario Goal (\%)}} &
        \multicolumn{2}{c}{\textbf{Avg.\ Interactions}} \\[0.15em]
        \cmidrule(lr){2-3}\cmidrule(lr){4-5}\cmidrule(lr){6-7}
         & Normal & Chall. & Normal & Chall. & Normal & Chall. \\
        \midrule
        All     & 73.2 & 57.6 & 62.5 & 48.2 & 10.69 & 8.40 \\
        Level 1 & 91.2 & 91.7 & 84.2 & 87.5 &  5.94 & 4.65 \\
        Level 2 & 77.1 & 58.7 & 68.8 & 42.0 & 10.36 & 8.33 \\
        Level 3 & 54.0 & 44.1 & 38.1 & 38.5 & 12.69 & 11.86 \\
        \bottomrule
    \end{tabular}}
    \label{tab:appworld}
\end{table}

These results validate the design of CUGA as a generalist computer-using agent: once the APIs and tools available to each sub-agent are defined, the planner–executor architecture can be configured to operate across domains without task-specific re-engineering. In practice, enterprises need only onboard their APIs and specify governance constraints for CUGA to extend its capabilities to new workflows.

\subsection{Enterprise Pilot: BPO-TA}

While CUGAs often perform well on abstract research benchmarks, enterprise deployment requires systematic offline evaluation with realistic tasks, audit guarantees, and compliance controls. To address this, we developed \textbf{BPO-TA}, a domain benchmark centered on talent acquisition (TA) workflows in the BPO context. BPO-TA encodes decision-support tasks drawn directly from analyst practice into a fixed test set, providing a reproducible regression baseline for measuring progress and ensuring operational reliability. Its design follows three principles essential for adoption: \emph{traceability} (each task paired with APIs, glue code, and gold-standard explanations), \emph{realism} (tasks grounded in genuine analyst workflows rather than synthetic probes), and \emph{reproducibility} (fixed inputs, deterministic evaluation, and explicit provenance).

The benchmark spans 26 tasks over 13 read-only APIs, covering endpoints such as \emph{SLA by source}, \emph{funnel conversion}, \emph{hires by source}, \emph{skill-impact on SLA}, \emph{definitions/methodology}, \emph{dataset/model lineage}, and \emph{timeframe metadata}. All endpoints are onboarded through the API/Tool Hub with minimized OpenAPI specs, validators, and read-only wrappers that strip or redact PII. Schema-grounded prompts enforce canonical definitions (e.g., SLA), and deterministic parsers gate LLM outputs. Every response includes a provenance panel listing API paths, parameters, and a computation log, enabling audit and regression testing. Operations are monitored for \emph{latency}, \emph{cost}, and \emph{policy compliance}.

Task categories mirror enterprise usage: (1) \emph{simple lookups} (e.g., requisition definitions), (2) \emph{cross-API joins} (e.g., linking candidate volume to conversion-to-hire), (3) \emph{looped reasoning} (e.g., filtering skills that negatively affect SLA), (4) \emph{provenance explanations} (e.g., surfacing dataset/model lineage), and (5) \emph{graceful failure}, where unsupported queries must be declined without hallucination. These patterns ensure that BPO-TA evaluates not only retrieval accuracy but also compositional reasoning, transparency, and robustness--making it both a research benchmark and an operational safeguard for trustworthy enterprise deployment.


\begin{table}[ht!]
    \centering
    \caption{Performance of CUGA on the \textbf{BPO-TA} benchmark.}
    \begin{tabular}{l c}
        \toprule
        \textbf{Metric} & \textbf{Value} \\
        \midrule
        Task Accuracy (26 tasks)        & \textbf{87\%} \\
        Valid First-Try Rate            & 78\% \\
        Responses with Provenance Logs  & 95\% \\
        Average Latency per Query       & 11.2s \\
        Analyst-Reported Reproducibility & 4.6 / 5 \\
        \bottomrule
    \end{tabular}
    \label{tab:bpota-results}
\end{table}

\paragraph{Results.} 
On the BPO-TA benchmark, CUGA achieves \textbf{87\% accuracy}, with failures concentrated on unsupported cross-application queries where graceful degradation is expected.  
Valid-first-try rates improved from 62\% (vanilla ReAct baseline) to 79\% with full CUGA.  
Ablations highlight the importance of reflective retries (-11 points without) and variable tracking (-15 reproducibility without).

\subsection{Potential Benefits}
The CUGA system has been piloted within IBM's Business-Process-Outsourcing (BPO) talent acquisition workflow since mid-2025. It can be used by recruiters and analyst teams to answer sourcing, funnel, and skill-impact questions that previously required manual data pulls and spreadsheet manipulation. The pilot was performed in a read-only configuration: CUGA connects to 13 domain-specific analytics APIs, each exposing pre-approved metrics such as funnel conversions and hires-by-source. Provenance logging and computation traces are stored for each interaction, ensuring audit readiness and compliance with organizational governance requirements (e.g., PII avoidance, immutable records of all API calls).

Preliminary evaluations of CUGA in simulated enterprise workflows, although not formally tested for statistical significance \citep{dror2018hitchhiker,dror2020statistical}, suggest promising efficiency and reliability gains.  suggest promising efficiency and reliability gains. Estimated benefits include a potential reduction in average time-to-answer (from roughly 20 minutes of manual work to an expected 2–5 minutes with CUGA, an \textit{estimated} $\sim$90 \% improvement) and higher reproducibility of responses (CUGA outputs were consistent across runs in about 90 \% of internal test cases). Audit readiness is also expected to improve, with over 90 \% of generated responses including full provenance (API endpoint, parameters, and result logs).

\begin{table}[ht]
\centering
\caption{Preliminary evaluation of estimated benefits in CUGA’s simulated enterprise use case (Talent Acquisition domain).}
\vspace{0.3em}
\resizebox{\linewidth}{!}{
\begin{tabular}{l c c}
\toprule
\textbf{Metric} & \textbf{Manual Baseline} & \textbf{CUGA (Pilot Evaluation)} \\
\midrule
Average time-to-answer & $\sim$20 min manual analysis & $\sim$2--5 min \\
Reproducibility of answers & $\sim$60\% & $\sim$95\% (test runs) \\
Responses with full provenance & $\sim$40\% & $\sim$92\% (expected) \\
Analyst effort (manual steps) & High (spreadsheets, queries) & Low (single agent call) \\
\midrule
\textit{Case study: skill impact analysis} & $\sim$30 min(manual SLA comparisons) & $\sim$6 min (projected) \\
\bottomrule
\end{tabular}}
\label{tab:deployment-benefits}
\end{table}

\noindent
Table~\ref{tab:deployment-benefits} summarizes these preliminary, pilot-level results from CUGA’s evaluation in the Talent Acquisition context. 
While the figures are based on controlled test environments and limited analyst feedback rather than full production deployment, 
they highlight the potential of generalist agents to enable substantial efficiency and transparency improvements in enterprise workflows.

\noindent
Qualitatively, BPO architects noted that CUGA can reduce reliance on ad hoc spreadsheet analysis, provide consistent explanations of sourcing and skill-impact decisions, and support faster onboarding for new team members through step-by-step reasoning grounded in enterprise APIs. 
Taken together, these preliminary observations indicate that generalist agents such as CUGA hold promise for delivering trustworthy, auditable, and scalable value as they transition from research prototypes toward enterprise-ready systems.

\subsection{Qualitative Evidence}
To complement quantitative metrics, we highlight two qualitative observations:

\paragraph{Case study.}
When asked \emph{“Which sourcing channel should we prioritize for requisition 05958BR?”}, CUGA queried two endpoints (\texttt{candidate\_volume}, \texttt{recommendation\_summary}), joined on source IDs, and produced a ranked table with SLA metrics. The interface presented not only the recommendation (“LinkedIn”) but also provenance: endpoint names, query parameters, and computation logs. Analysts reported this saved 20–30 minutes of manual dashboard comparisons.

\paragraph{Feedback.}
BPO architects emphasized reduced “spreadsheet wrangling,” describing CUGA as “freeing time for actual decision-making.” They noted that the ability to decline unsupported requests (e.g., region-level metrics not exposed by APIs) increased their trust, since the agent did not hallucinate unavailable data.


\section{Lessons Learned and Insights}
\label{sec:lessons}

The first pilot of CUGA in the BPO Talent Acquisition (TA) context provides early indications that generalist agents can move beyond benchmarks toward enterprise-grade applicability. 
Phase~1 of the application focused on use cases such as automated candidate scheduling and communication, pipeline visibility, and smart sourcing suggestions. 
Based on internal projections and controlled simulations, this approach may enable approximately 35\% of candidate inquiries to be resolved via self-service and 25\% recruiter workflow automation, alongside an estimated 90\% reduction in development time and a 50\% reduction in development cost compared to task-specific baselines. 
These preliminary outcomes suggest that generalist agents have the potential to accelerate time-to-value while maintaining the governance and transparency required in enterprise workflows.

From this pilot, we derived a set of technical and organizational lessons that shape the path forward:

\paragraph{Technical insights.}
\begin{itemize}
    \item \textbf{Prompt and specification curation.} Minimizing OpenAPI specifications and keeping prompts concise, schema-grounded, and unambiguous substantially improved reliability and efficiency.
    \item \textbf{Governance alignment.} Restricting to read-only APIs, redacting personally identifiable information, and grounding all answers in canonical definitions were essential to gain organizational trust.
    \item \textbf{Reliability mechanisms.} Interrupt nodes, reflective retries, and a code planner (generating structured pseudo-code) reduced failure rates and improved reproducibility.
    \item \textbf{Monitoring and reproducibility.} Provenance logs and regression testing enabled both audit readiness and systematic debugging of failure modes.
    \item \textbf{Sustainability and extensibility.} The API/Tool Hub streamlined onboarding of new endpoints, allowing rapid iteration as business requirements evolved.
    \item \textbf{Analytical foundation.} A critical advantage was the architecture’s ability to log, monitor, and analyze agent decisions. This enabled systematic investigation of failures and deeper insight into why the agent behaved as it did, laying the groundwork for continuous improvement.
\end{itemize}

\paragraph{Organizational insights.}
\begin{itemize}
    \item \textbf{Human-in-the-loop configuration.} Business users required explicit control over when the agent could act autonomously versus when approval was mandatory, making configurable HITL a central requirement.
    \item \textbf{Benchmarks are not enough.} While AppWorld and WebArena validated general capabilities, enterprise adoption depended on a domain-specific benchmark (BPO-TA) that reflected real recruiter workflows and enabled regression testing.
    \item \textbf{Bridging research and operations.} Deployment success depended as much on organizational alignment--policies, governance, and HITL practices--as on technical breakthroughs. The transition from a promising demo to a trusted production system required deliberate discipline and collaboration across business and research teams.
\end{itemize}

CUGA’s first phase surfaced clear requirements for enterprise deployment, driving current work on configurable human-in-the-loop control, explicit policy-enforcement for safe autonomous actions, improved cost--latency tradeoffs through adaptive short-circuiting, reuse of successful trajectories as tools, and selective use of smaller models for routine tasks. The next milestone is evaluation on policy compliance and HITL governance as an enterprise-ready system that meets organizational standards of safety and trust.




\section{Conclusion}
\label{sec:conclusion}
This work provides early evidence that generalist agents can enable measurable business value in enterprise contexts. By combining layered planning, provenance-aware execution, and governance alignment, CUGA can reduce time-to-answer, improve reproducibility, and enable trustworthy automation in talent acquisition. The lessons from Phase~1--both technical and organizational--show that moving from research breakthroughs to enterprise readiness is less about a single algorithmic leap and more about disciplined engineering, governance, and continuous iteration. Architectures like CUGA mark a credible path toward enterprise-ready generalist agents that are safe, efficient, and adaptable.

\bibliography{aaai2026}

@inproceedings{zhou2023webarena,
  title={WebArena: A Realistic Web Environment for Building Autonomous Agents},
  author={Zhou, Shuyan and Xu, Frank F and Zhu, Hao and Zhou, Xuhui and Lo, Robert and Sridhar, Abishek and Cheng, Xianyi and Ou, Tianyue and Bisk, Yonatan and Fried, Daniel and others},
  booktitle={The Twelfth International Conference on Learning Representations},
  year={2023}
}

@inproceedings{
xie2024osworld,
title={{OSW}orld: Benchmarking Multimodal Agents for Open-Ended Tasks in Real Computer Environments},
author={Tianbao Xie and Danyang Zhang and Jixuan Chen and Xiaochuan Li and Siheng Zhao and Ruisheng Cao and Toh Jing Hua and Zhoujun Cheng and Dongchan Shin and Fangyu Lei and Yitao Liu and Yiheng Xu and Shuyan Zhou and Silvio Savarese and Caiming Xiong and Victor Zhong and Tao Yu},
booktitle={The Thirty-eight Conference on Neural Information Processing Systems Datasets and Benchmarks Track},
year={2024},
url={https://openreview.net/forum?id=tN61DTr4Ed}
}

@article{shinn2023reflexion,
  title={Reflexion: Language agents with verbal reinforcement learning},
  author={Shinn, Noah and Cassano, Federico and Gopinath, Ashwin and Narasimhan, Karthik and Yao, Shunyu},
  journal={Advances in Neural Information Processing Systems},
  volume={36},
  pages={8634--8652},
  year={2023}
}

@article{nakano2021webgpt,
  title={WebGPT: Browser-assisted question-answering with human feedback},
  author={Reiichiro Nakano and Jacob Hilton and Suchir Balaji and Jeff Wu and Ouyang Long and Christina Kim and Christopher Hesse and Shantanu Jain and Vineet Kosaraju and William Saunders and Xu Jiang and Karl Cobbe and Tyna Eloundou and Gretchen Krueger and Kevin Button and Matthew Knight and Benjamin Chess and John Schulman},
  journal={ArXiv},
  year={2021},
  volume={abs/2112.09332},
  url={https://api.semanticscholar.org/CorpusID:245329531}
}

@misc{openai2025operator,
  title={Introducing Operator},
  author={{OpenAI}},
  year={2025},
  howpublished={\url{https://openai.com/index/introducing-operator/}},
  note={Accessed: 2025-08-14}
}

@misc{anthropic2024computeruse,
  title={Introducing computer use, a new Claude 3.5 Sonnet, and more},
  author={{Anthropic}},
  year={2024},
  howpublished={\url{https://www.anthropic.com/news/3-5-models-and-computer-use}},
  note={Accessed: 2025-08-14}
}

@inproceedings{raghavan2020mitigating,
  title={Mitigating bias in algorithmic hiring: Evaluating claims and practices},
  author={Raghavan, Manish and Barocas, Solon and Kleinberg, Jon and Levy, Karen},
  booktitle={Proceedings of the 2020 Conference on Fairness, Accountability, and Transparency},
  pages={469--481},
  year={2020}
}

@misc{nist2023airmf,
  author = {Elham Tabassi},
  title = {Artificial Intelligence Risk Management Framework (AI RMF 1.0)},
  year = {2023},
  month = {2023-01-26 05:01:00},
  publisher = {NIST Trustworthy and Responsible AI, National Institute of Standards and Technology, Gaithersburg, MD},
  url = {https://tsapps.nist.gov/publication/get_pdf.cfm?pub_id=936225},
  doi = {https://doi.org/10.6028/NIST.AI.100-1},
  language = {en}
}

@misc{euai2024,
  title={Regulation (EU) 2024/1689 of the European Parliament and of the Council of 13 June 2024 laying down harmonised rules on artificial intelligence (Artificial Intelligence Act)},
  author={{European Union}},
  year={2024},
  howpublished={\url{https://eur-lex.europa.eu/eli/reg/2024/1689/oj/eng}},
  note={OJ L 2024/1689, 12.7.2024}
}

@misc{nyc144,
  title={{NYC Local Law 144} and Final Rules on Automated Employment Decision Tools},
  author={{NYC Department of Consumer and Worker Protection}},
  year={2023},
  howpublished={\url{https://www.nyc.gov/assets/dca/downloads/pdf/rules/Rules-Amendment-6RCNY5-300-AEDT.pdf}},
  note={Accessed 2025-08-19}
}

@article{fabris2025fairness,
  title={Fairness and Bias in Algorithmic Hiring: A Multidisciplinary Survey},
  author={Fabris, Alessandro and Baranowska, Nina and Dennis, Matthew J and Graus, David and Hacker, Philipp and Saldivar, Jorge and Zuiderveen Borgesius, Frederik and Biega, Asia J},
  journal={ACM Transactions on Intelligent Systems and Technology},
  year={2025}
}

@article{chen2023ethics,
  title={Ethics and discrimination in artificial intelligence-enabled recruitment practices},
  author={Chen, Zhisheng},
  journal={Humanities and Social Sciences Communications},
  volume={10},
  number={1},
  pages={1--13},
  year={2023},
  publisher={Nature Portfolio}
}

@inproceedings{yao2023react,
  title={React: Synergizing reasoning and acting in language models},
  author={Yao, Shunyu and Zhao, Jeffrey and Yu, Dian and Du, Nan and Shafran, Izhak and Narasimhan, Karthik R and Cao, Yuan},
  booktitle={The eleventh international conference on learning representations},
  year={2022}
}

@inproceedings{ye2024codeact,
  title={Executable code actions elicit better llm agents},
  author={Wang, Xingyao and Chen, Yangyi and Yuan, Lifan and Zhang, Yizhe and Li, Yunzhu and Peng, Hao and Ji, Heng},
  booktitle={Forty-first International Conference on Machine Learning},
  year={2024}
}

@inproceedings{wu2023autogen,
  title={Autogen: Enabling next-gen LLM applications via multi-agent conversations},
  author={Wu, Qingyun and Bansal, Gagan and Zhang, Jieyu and Wu, Yiran and Li, Beibin and Zhu, Erkang and Jiang, Li and Zhang, Xiaoyun and Zhang, Shaokun and Liu, Jiale and others},
  booktitle={First Conference on Language Modeling},
  year={2024}
}

@online{langgraph2024,
  title        = {LangGraph Documentation: Building Stateful, Multi-Agent Workflows},
  author       = {{LangChain}},
  year         = {2024},
  url          = {https://langchain-ai.github.io/langgraph/},
  note         = {Accessed 19-Aug-2025}
}

@online{openaiSwarm2024,
  title        = {OpenAI Swarm: Lightweight Multi-Agent Orchestrator},
  author       = {{OpenAI}},
  year         = {2024},
  url          = {https://github.com/openai/swarm},
  note         = {Accessed 19-Aug-2025}
}

@inproceedings{trivedi2024appworld,
  title        = {AppWorld: A Controllable World of Apps and People for Benchmarking Interactive Coding Agents},
  author       = {Trivedi, Harsh and Khot, Tushar and Hartmann, Mareike and Manku, Ruskin and Dong, Vinty and Li, Edward and Gupta, Shashank and Sabharwal, Ashish and Balasubramanian, Niranjan},
  booktitle    = {Proceedings of the 62nd Annual Meeting of the Association for Computational Linguistics (ACL)},
  year         = {2024},
  url          = {https://aclanthology.org/2024.acl-long.850.pdf}
}

@inproceedings{jiang2023llmlingua,
  title={LLMLingua: Compressing Prompts for Accelerated Inference of Large Language Models},
  author={Jiang, Huiqiang and Wu, Qianhui and Lin, Chin-Yew and Yang, Yuqing and Qiu, Lili},
  booktitle={The 2023 Conference on Empirical Methods in Natural Language Processing},
year = {2023}
}

@online{verizonVerge2025,
  title        = {Verizon adopts Google's Gemini AI to help customers solve `complex' issues},
  author       = {{The Verge}},
  year         = {2025},
  month        = {June},
  url          = {https://www.theverge.com/news/691810/verizon-google-gemini-ai-chatbot-customer-service},
  note         = {Accessed 19-Aug-2025}
}

@online{verizonMWL2025,
  title        = {Verizon lauds Google Cloud AI customer service move},
  author       = {{Mobile World Live}},
  year         = {2025},
  month        = {April},
  url          = {https://www.mobileworldlive.com/verizon/verizon-lauds-google-cloud-ai-customer-service-move/},
  note         = {Accessed 19-Aug-2025}
}

@online{aibusinessAlibaba2022,
  title        = {Alibaba turns to AI to cut customer service costs},
  author       = {{AI Business}},
  year         = {2022},
  month        = {October},
  url          = {https://aibusiness.com},
  note         = {Coverage of large-scale customer service automation; Accessed 19-Aug-2025}
}

@online{snowflakeCortex2024,
  title        = {Introducing Snowflake Cortex Analyst: Natural Language to Accurate SQL},
  author       = {{Snowflake Engineering}},
  year         = {2024},
  url          = {https://www.snowflake.com/blog/snowflake-cortex-analyst/},
  note         = {Accessed 19-Aug-2025}
}

@online{msPowerBI2024,
  title        = {Power BI Copilot is now generally available},
  author       = {{Microsoft}},
  year         = {2024},
  url          = {https://techcommunity.microsoft.com/t5/power-bi-blog/copilot-in-power-bi-is-now-generally-available/ba-p/4149995},
  note         = {Accessed 19-Aug-2025}
}

@online{databricksGenie2025,
  title        = {Introducing Genie: AI/BI for the Lakehouse},
  author       = {{Databricks}},
  year         = {2025},
  month        = {April},
  url          = {https://www.databricks.com/blog/introducing-genie-ai-bi},
  note         = {Accessed 19-Aug-2025}
}

@article{levy2025stwebagentbench,
  title={St-webagentbench: A benchmark for evaluating safety and trustworthiness in web agents},
  author={Levy, Ido and Wiesel, Ben and Marreed, Sami and Oved, Alon and Yaeli, Avi and Shlomov, Segev},
  journal={arXiv preprint arXiv:2410.06703},
  year={2024}
}

@article{shen2025mind,
  title={From mind to machine: The rise of manus ai as a fully autonomous digital agent},
  author={Shen, Minjie and Li, Yanshu and Chen, Lulu and Yang, Qikai},
  journal={arXiv preprint arXiv:2505.02024},
  year={2025}
}

@article{fourney2024magentic,
  title={Magentic-one: A generalist multi-agent system for solving complex tasks},
  author={Fourney, Adam and Bansal, Gagan and Mozannar, Hussein and Tan, Cheng and Salinas, Eduardo and Niedtner, Friederike and Proebsting, Grace and Bassman, Griffin and Gerrits, Jack and Alber, Jacob and others},
  journal={arXiv preprint arXiv:2411.04468},
  year={2024}
}

@article{zhang2025agentorchestra,
  title={Agentorchestra: A hierarchical multi-agent framework for general-purpose task solving},
  author={Zhang, Wentao and Cui, Ce and Zhao, Yilei and Hu, Rui and Liu, Yang and Zhou, Yahui and An, Bo},
  journal={arXiv preprint arXiv:2506.12508},
  year={2025}
}

@article{marreed2025towards,
  title={Towards enterprise-ready computer using generalist agent},
  author={Marreed, Sami and Oved, Alon and Yaeli, Avi and Shlomov, Segev and Levy, Ido and Akrabi, Offer and Sela, Aviad and Adi, Asaf and Mashkif, Nir},
  journal={arXiv preprint arXiv:2503.01861},
  year={2025}
}

@inproceedings{yao2024tau,
  title={Tau-bench: A Benchmark for Tool-Agent-User Interaction in Real-World Domains},
  author={Yao, Shunyu and Shinn, Noah and Razavi, Pedram and Narasimhan, Karthik R},
  booktitle={The Thirteenth International Conference on Learning Representations},
  year={2025}
}

@article{debrowsergym,
  title={The BrowserGym Ecosystem for Web Agent Research},
  author={de Chezelles, Thibault Le Sellier and Gasse, Maxime and Lacoste, Alexandre and Caccia, Massimo and Drouin, Alexandre and Boisvert, L{\'e}o and Thakkar, Megh and Marty, Tom and Assouel, Rim and Shayegan, Sahar Omidi and others},
  journal={Transactions on Machine Learning Research},
  year={2024}
}

@article{kim2023language,
  title={Language models can solve computer tasks},
  author={Kim, Geunwoo and Baldi, Pierre and McAleer, Stephen},
  journal={Advances in Neural Information Processing Systems},
  volume={36},
  pages={39648--39677},
  year={2023}
}

@article{gupta2025leveraging,
  title={Leveraging In-Context Learning for Language Model Agents},
  author={Gupta, Shivanshu and Singh, Sameer and Sabharwal, Ashish and Khot, Tushar and Bogin, Ben},
  journal={arXiv preprint arXiv:2506.13109},
  year={2025}
}

@article{chen2025reinforcement,
  title={Reinforcement learning for long-horizon interactive llm agents},
  author={Chen, Kevin and Cusumano-Towner, Marco and Huval, Brody and Petrenko, Aleksei and Hamburger, Jackson and Koltun, Vladlen and Kr{\"a}henb{\"u}hl, Philipp},
  journal={arXiv preprint arXiv:2502.01600},
  year={2025}
}

@inproceedings{zhang2025agent,
  title={Which Agent Causes Task Failures and When? On Automated Failure Attribution of LLM Multi-Agent Systems},
  author={Zhang, Shaokun and Yin, Ming and Zhang, Jieyu and Liu, Jiale and Han, Zhiguang and Zhang, Jingyang and Li, Beibin and Wang, Chi and Wang, Huazheng and Chen, Yiran and others},
  booktitle={Forty-second International Conference on Machine Learning},
  year={2025}
}

@inproceedings{cemri2025multi,
  title={Why do multiagent systems fail?},
  author={Pan, Melissa Z and Cemri, Mert and Agrawal, Lakshya A and Yang, Shuyi and Chopra, Bhavya and Tiwari, Rishabh and Keutzer, Kurt and Parameswaran, Aditya and Ramchandran, Kannan and Klein, Dan and others},
  booktitle={ICLR 2025 Workshop on Building Trust in Language Models and Applications},
  year={2025}
}

@inproceedings{shen2024scribeagent,
  title={WorkflowAgent: Towards Specialized Web Agents Using Production-Scale Workflow Data},
  author={Shen, Junhong and Jain, Atishay and Xiao, Zedian and Amlekar, Ishan and Hadji, Mouad and Podolny, Aaron and Talwalkar, Ameet},
  booktitle={ICLR 2025 Workshop on Foundation Models in the Wild},
  year={2024}
}

@article{zhang2025symbiotic,
  title={Symbiotic cooperation for web agents: Harnessing complementary strengths of large and small llms},
  author={Zhang, Ruichen and Qiu, Mufan and Tan, Zhen and Zhang, Mohan and Lu, Vincent and Peng, Jie and Xu, Kaidi and Agudelo, Leandro Z and Qian, Peter and Chen, Tianlong},
  journal={arXiv preprint arXiv:2502.07942},
  year={2025}
}

@misc{jaceAIWebArenaResults,
	author = {Jace.AI},
	title = {jace.ai},
	howpublished = {\url{https://jace.ai/blog/awa-1-5-achieves-breakthrough-performance-on-web-arena-benchmark}},
	year = {2024},
	note = {},
}

@article{shen2024taskbench,
  title={Taskbench: Benchmarking large language models for task automation},
  author={Shen, Yongliang and Song, Kaitao and Tan, Xu and Zhang, Wenqi and Ren, Kan and Yuan, Siyu and Lu, Weiming and Li, Dongsheng and Zhuang, Yueting},
  journal={Advances in Neural Information Processing Systems},
  volume={37},
  pages={4540--4574},
  year={2024}
}

@inproceedings{xu2023tool,
  title={On the Tool Manipulation Capability of Open-sourced Large Language Models},
  author={Xu, Qiantong and Hong, Fenglu and Li, Bo and Hu, Changran and Chen, Zhengyu and Zhang, Jian},
  booktitle={NeurIPS 2023 Foundation Models for Decision Making Workshop},
      year={2023},
}

@article{zheng2024sglang,
  title={Sglang: Efficient execution of structured language model programs},
  author={Zheng, Lianmin and Yin, Liangsheng and Xie, Zhiqiang and Sun, Chuyue Livia and Huang, Jeff and Yu, Cody Hao and Cao, Shiyi and Kozyrakis, Christos and Stoica, Ion and Gonzalez, Joseph E and others},
  journal={Advances in neural information processing systems},
  volume={37},
  pages={62557--62583},
  year={2024}
}

@inproceedings{kwon2023efficient,
  title={Efficient memory management for large language model serving with pagedattention},
  author={Kwon, Woosuk and Li, Zhuohan and Zhuang, Siyuan and Sheng, Ying and Zheng, Lianmin and Yu, Cody Hao and Gonzalez, Joseph and Zhang, Hao and Stoica, Ion},
  booktitle={Proceedings of the 29th symposium on operating systems principles},
  pages={611--626},
  year={2023}
}

@inproceedings{oved2025snap,
  title={SNAP: semantic stories for next activity prediction},
  author={Oved, Alon and Shlomov, Segev and Zeltyn, Sergey and Mashkif, Nir and Yaeli, Avi},
  booktitle={Proceedings of the AAAI Conference on Artificial Intelligence},
  volume={39},
  pages={28871--28877},
  year={2025}
}

@article{shlomov2024grounding,
  title={From grounding to planning: Benchmarking bottlenecks in web agents},
  author={Shlomov, Segev and Sela, Aviad and Levy, Ido and Galanti, Liane and Abitbol, Roy and others},
  journal={arXiv preprint arXiv:2409.01927},
  year={2024}
}

@article{shlomov2024ida,
  title={Ida: Breaking barriers in no-code ui automation through large language models and human-centric design},
  author={Shlomov, Segev and Yaeli, Avi and Marreed, Sami and Schwartz, Sivan and Eder, Netanel and Akrabi, Offer and Zeltyn, Sergey},
  journal={arXiv preprint arXiv:2407.15673},
  year={2024}
}

@inproceedings{yaeli2022recommending,
  title={Recommending next best skill in conversational robotic process automation},
  author={Yaeli, Avi and Shlomov, Segev and Oved, Alon and Zeltyn, Sergey and Mashkif, Nir},
  booktitle={International Conference on Business Process Management},
  pages={215--230},
  year={2022},
  organization={Springer}
}

@book{dror2020statistical,
  title={Statistical significance testing for natural language processing},
  author={Dror, Rotem and Peled-Cohen, Lotem and Shlomov, Segev and Reichart, Roi},
  year={2020},
  publisher={Springer}
}

@article{zeltyn2022prescriptive,
  title={Prescriptive process monitoring in intelligent process automation with chatbot orchestration},
  author={Zeltyn, Sergey and Shlomov, Segev and Yaeli, Avi and Oved, Alon},
  journal={arXiv preprint arXiv:2212.06564},
  year={2022}
}

@article{schwartz2023enhancing,
  title={Enhancing trust in LLM-based AI automation agents: New considerations and future challenges},
  author={Schwartz, Sivan and Yaeli, Avi and Shlomov, Segev},
  journal={arXiv preprint arXiv:2308.05391},
  year={2023}
}

@inproceedings{dror2018hitchhiker,
  title={The hitchhiker’s guide to testing statistical significance in natural language processing},
  author={Dror, Rotem and Baumer, Gili and Shlomov, Segev and Reichart, Roi},
  booktitle={Proceedings of the 56th Annual Meeting of the Association for Computational Linguistics (Volume 1: Long Papers)},
  pages={1383--1392},
  year={2018}
}

\newpage
\appendix
\clearpage

\section{CUGA System Architecture}
\label{app:architecture}

This section presents the detailed architecture of the CUGA system, including both the Browser and API sub-agents.

\begin{figure}[ht!]
    \centering
    \includegraphics[width=0.8\linewidth]{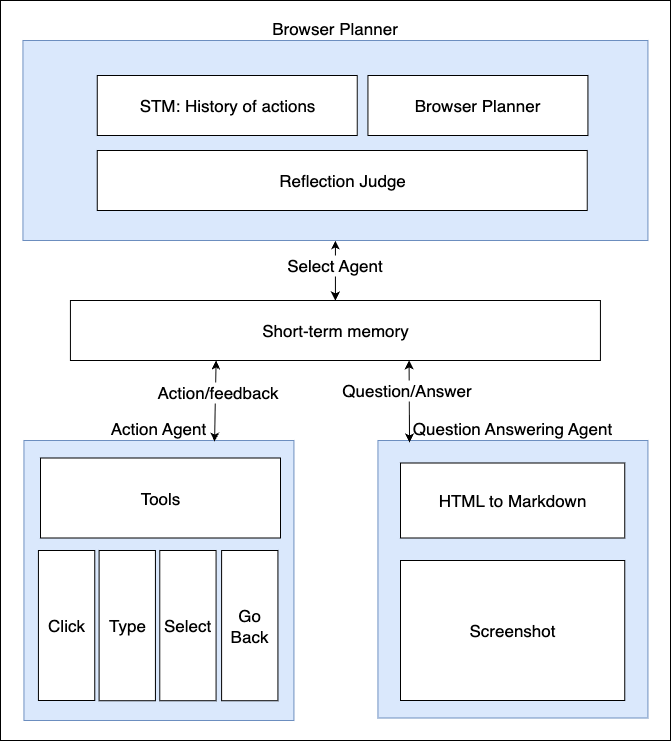}
    \caption{Architecture of the Browser Sub Agent: the Browser Planner combines action history and reflection to select between two execution paths--an Action Agent (click, type, select, navigate) and a Question Answering Agent (DOM-to-Markdown conversion and screenshots).}
    \label{fig:browser-arch}
\end{figure}

\begin{figure}[ht!]
    \centering
    \includegraphics[width=0.8\linewidth]{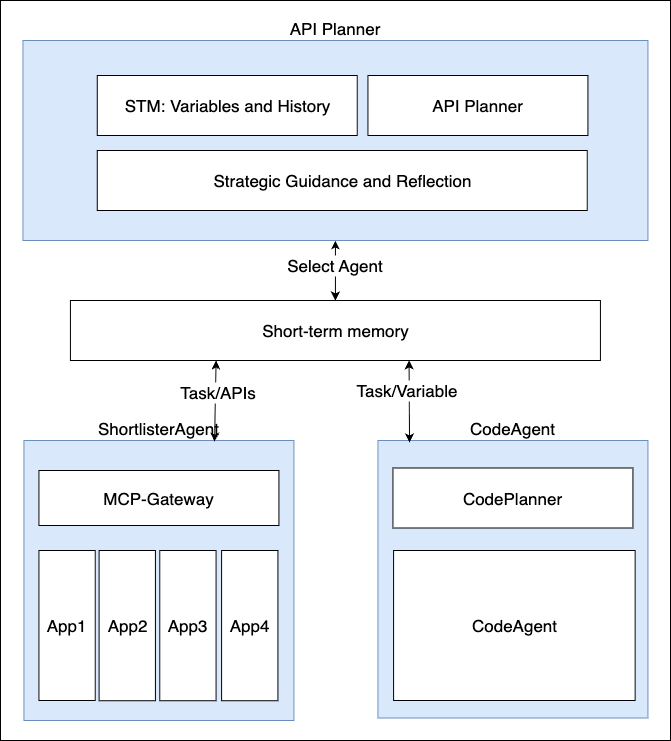}
    \caption{Architecture of the API Sub Agent: the API Planner coordinates short-term memory, reflection, and two executors--the ShortlisterAgent, which routes to APIs through a registry, and the CodeAgent, which generates and executes code via a nested CodePlanner.}
    \label{fig:api-arch}
\end{figure}

\section{CUGA's Nodes}

\subsection*{ChatAgent | Chat}
This agent manages follow-up questions and handles conversations that trigger pre-existing, familiar flows.

\subsection*{TaskAnalyzerAgent | Task Analyzer}
This agent performs the initial analysis of a user's request to determine its complexity. It decides whether a simple, direct answer is sufficient or if the query requires a more complex, multi-step plan. This agent also identifies related applications based on the user's intent.

\subsection*{TaskDecompositionAgent | Task Decomposition}
The agent decomposes complex tasks into smaller, manageable sub-tasks and assigns each to a specialized agent. This approach is essential for effectively solving multi-step problems, as it allows each sub-task to be carefully planned and executed..

\subsection*{PlanControllerAgent | Plan Controller}
This agent acts as the central orchestrator of the sub-agents. It reviews the overall plan, tracks the status of sub-tasks created by \textbf{TaskDecompositionAgent}, and decides the next steps, ensuring the entire process stays on track from decomposition to the final answer.

\subsection*{Browser Sub-Agent}

\subsubsection*{BrowserPlannerAgent | Browser Planner}
This agent plans the next steps in natural language, based on its understanding of the DOM page and accompanying images. It then passes tasks to either the \textbf{ActionAgent} or \textbf{QaAgent}, or decides to conclude the task.

\subsubsection*{ActionAgent | Action Agent}
This agent is designed to perform specific subtasks directly on the current web page, such as clicking on an element.

\subsubsection*{QaAgent | Question Answering Agent}
This agent is called upon to answer questions related to the current webpage.

\section*{Human-in-the-Loop Nodes}

\subsection*{SuggestHumanActions | Human Action Suggester}
This node is used when human intervention or input would be beneficial. It's necessary for collaborative workflows, allowing the AI to ask for help, clarification, or a decision from the user.

\subsection*{WaitForResponse | Response Waiter}
This node pauses the entire workflow until it receives a response, typically from a human user. It works in conjunction with \textbf{SuggestHumanActions} to enable true human-in-the-loop processing.

\section*{API Sub-Agent}

\subsection*{APIPlannerAgent | API Planner}
This agent specializes in creating sub-tasks that involve API calls. At each turn, it either calls the \textbf{ShortlisterAgent} or \textbf{APICodePlannerAgent}, or decides to finish the task and return to the \textbf{PlanControllerAgent}. Within this node, there is also a reflection component that runs upon returning from the \textbf{CodeAgent}, summarizing the task, checking edge cases, and suggesting strategic recommendations.

\subsection*{APICodePlannerAgent | API Code Planner}
This agent, given the shortlisted schema and sub-task decided by the \textbf{APIPlannerAgent}, generates a pseudo-natural language plan to guide the coding agent. It can also report missing APIs to the \textbf{APIPlannerAgent} as feedback.

\subsection*{CodeAgent | Coding Agent}
This agent is responsible for writing and executing code to solve a problem. It is called after the \textbf{APICodePlannerAgent}. The generated code is executed in a sandbox to ensure safety, and the result is stored in a variable and fed back to the \textbf{APIPlannerAgent} concisely within its context.

\subsection*{ShortlisterAgent | Tool Shortlister}
This agent filters and ranks a list of available tools or APIs to find the most relevant ones for a given sub-task generated by the \textbf{APIPlannerAgent}.

\section*{Other Agents}

\subsection*{FinalAnswerAgent | Final Answer}
This agent is responsible for gathering the results from all completed tasks and synthesizing them into a final, coherent response for the user. It represents the last step in the cognitive process before presenting the solution.

\subsection*{ReuseAgent | Reuse Agent}
This agent is used in "save \& reuse" mode, where CUGA suggests that the user save the current autonomous flow into deterministic Python code for safer and more predictable execution. It runs after the \textbf{FinalAnswerAgent} in conjunction with human-in-the-loop actions.

\newpage
\section{WebArena Benchmark Analysis}
\label{dataset:webarena}
The WebArena \cite{zhou2023webarena} environment is a realistic and reproducible web environment designed for web agents to interact with. Along with the environment, a benchmark that consists of a set of 812 real-life natural language tasks has been released.

\begin{figure}[htbp]
    \centering
    \includegraphics[width=\columnwidth]{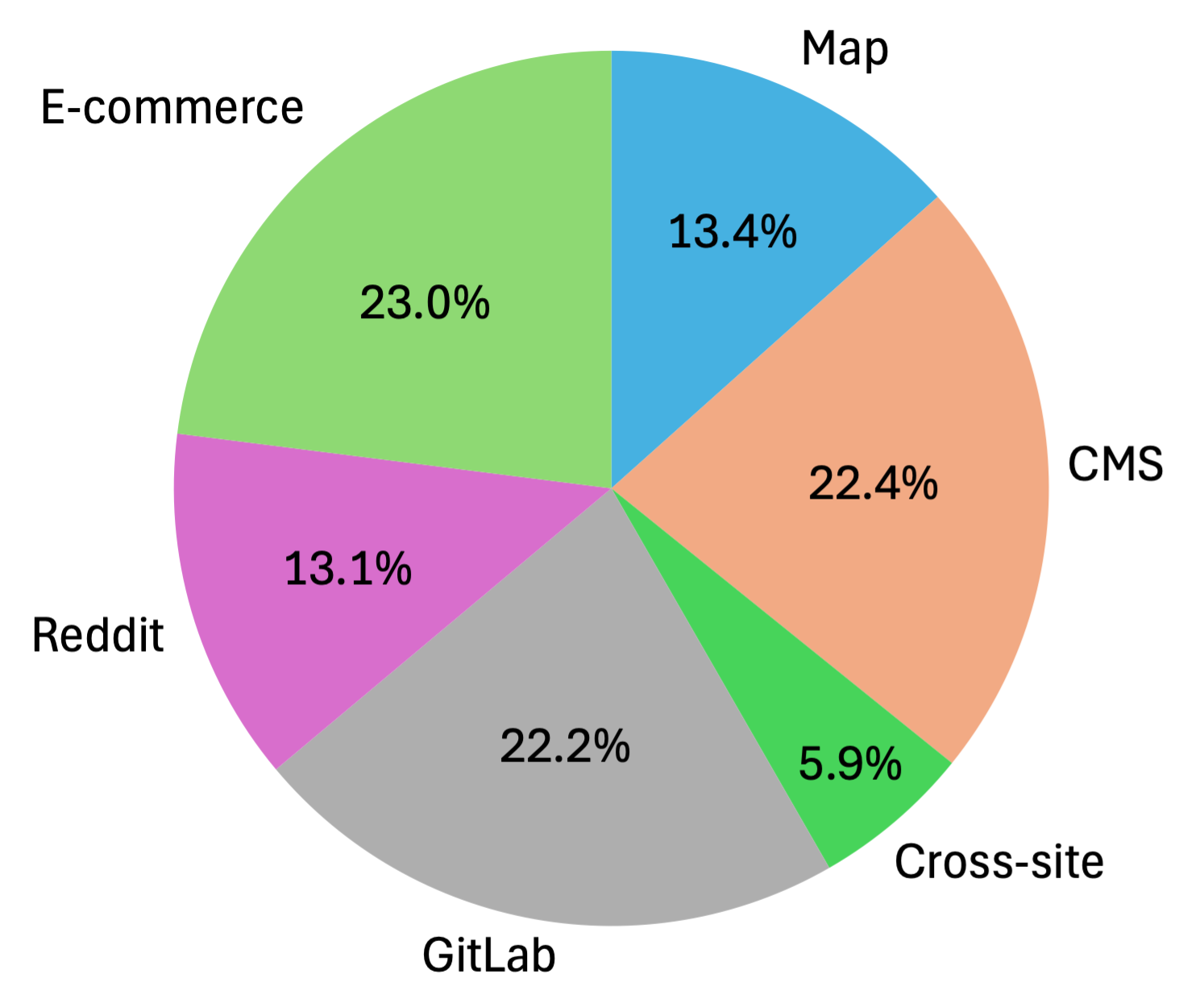}
    \caption{WebArena intent distribution across different websites. Cross-site intents include interaction with multiple websites.}
    \label{fig:webarena-composition}
\end{figure}

\begin{table}[htbp]
\centering
\label{tab:webarena_leaderboard}
\begin{tabular*}{\columnwidth}{l|c}
\hline
\textbf{Method} & \textbf{SR (\%)} \\
\hline
IBM CUGA (ours) & 61.7 \\
Operator \cite{openai2025operator} & 58.1 \\
Autonomous Web Agent \cite{jaceAIWebArenaResults} & 57.1 \\
ScribeAgent + GPT-4o \cite{shen2024scribeagent} & 53 \\
AgentSymbiotic \cite{zhang2025symbiotic} & 52.1 \\
\hline
\end{tabular*}
\caption{WebArena Benchmark Leaderboard. SR = Success Rate}
\end{table}

\begin{figure}[ht!]
    \centering
    \includegraphics[width=0.6\columnwidth]{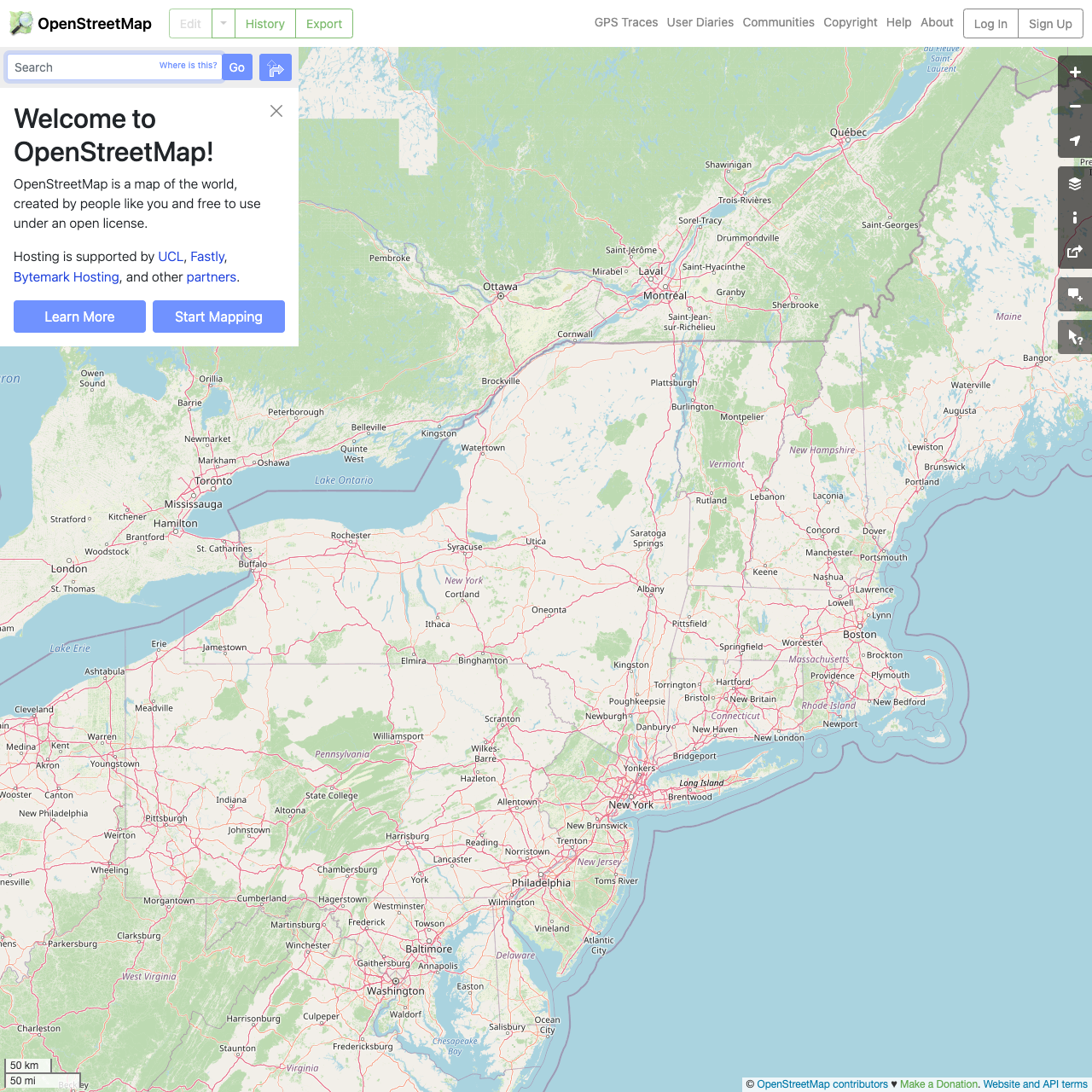}
    \caption{WebArena Map domain task example}
    \label{fig:webarena-map-task}
\end{figure}

\section{AppWorld Dataset}
\label{dataset:appworld}
AppWorld \cite{trivedi2024appworld} is a high-quality execution environment designed to test agents with code generation capabilities. The AppWorld benchmark consists of 9 day-to-day application such as Spotify, Amazon Shopping, Gmail, available via 457 APIs and simulating the lives of simulated users. The benchmark is divided into two levels of difficulty - named "Test-Normal" and "Test-Challenge", with the latter being generally more challenging, with each level consisting of several scenario data points, with each scenario consisting of several tasks for an agent to accomplish. 

The metrics used to test agents in this benchmark are \textbf{Task Goal Completion}, the percentage of tasks for which the agent passed all evaluation tests, and \textbf{Scenario Goal Completion}, the percentage of task \textit{scenarios} for which the agent passed all evaluation tests for \textit{all} tasks from that scenario.

\begin{table}[ht!]
\centering
\label{tab:appworld_composition}
\begin{tabular*}{\columnwidth}{l|c|c}
\hline
 & Test-Normal & Test-Challenge \\
 & avg (max) & avg (max) \\
\hline
Num. Apps & 1.5 (3) & 2.0 (6) \\
Num. Unique APIs & 8.2 (17) & 10.5 (26) \\
Num. API calls & 42.5 (244) & 46.8 (649) \\
Num. Sol. Code Lines & 41.3 (134) & 56.9 (128) \\
Num. Eval. Tests & 5.9 (19) & 8.0 (24) \\
Difficiculty Level & 1.9 (3) & 2.3 (3) \\
\hline
\end{tabular*}
\caption{AppWorld Benchmark Intent Composition}
\end{table}

\begin{table}[ht!]
\centering
\setlength{\tabcolsep}{1mm}
\label{tab:appworld_leaderboard}
\begin{tabular*}{\columnwidth}{l|c|c|c}
\hline
 & & Normal & Challenge \\
 Method & LLM & TGC (SGC) & TGC (SGC) \\
\hline
CUGA (ours) & GPT-4.1 & 73.2 (62.5) & 57.6 (48.2) \\
\citet{chen2025reinforcement} & Qwen2.5 & 72.6 (53.6) & 47.2 (28.8) \\
\citet{gupta2025leveraging} & GPT-4o & 68.5 (57.1) & 38.9 (23) \\
ReAct & GPT-4o & 48.8 (32.1) & 30.2 (13) \\
\hline
\end{tabular*}
\caption{AppWorld Benchmark Official Leaderboard. TGC and SGC metrics mean task and scenario goal completion respectively.}
\end{table}

\newpage
\section{BPO-TA Benchmark Details}
\label{app:bpota}

This section provides comprehensive details about the BPO-TA (Business Process Operations - Talent Acquisition) benchmark, including task taxonomy, catalog, and example queries.

\subsection{Task Categories and Examples}

\begin{table*}[ht!]
\centering\resizebox{1.0\linewidth}{!}{
\begin{tabular}{ll}
\toprule
\textbf{Category} & \textbf{Example Query} \\
\midrule
Lookup & ``How is the SLA metric defined for 05958BR?'' \\
Join & ``For requisitions like 05958BR, which sources provided the most candidates, and how effective were they at converting to hires?'' \\
Looped comparison & ``Out of Python, Quantum Physics, Cyber Engineering, Risk Analysis, Wireshark -- which skills negatively affect SLA?'' \\
Provenance explanation & ``What models and datasets were used to compute the SLA impact of Python?'' \\
Graceful failure & ``Who's the hiring manager for 05959BR and how responsive is she?'' \\
\bottomrule
\end{tabular}}
\caption{BPO-TA benchmark categories with representative queries.}
\label{tab:bpota}
\end{table*}

\subsection{Dataset Taxonomy}

\begin{figure*}[ht!]
    \centering
    \includegraphics[width=\linewidth]{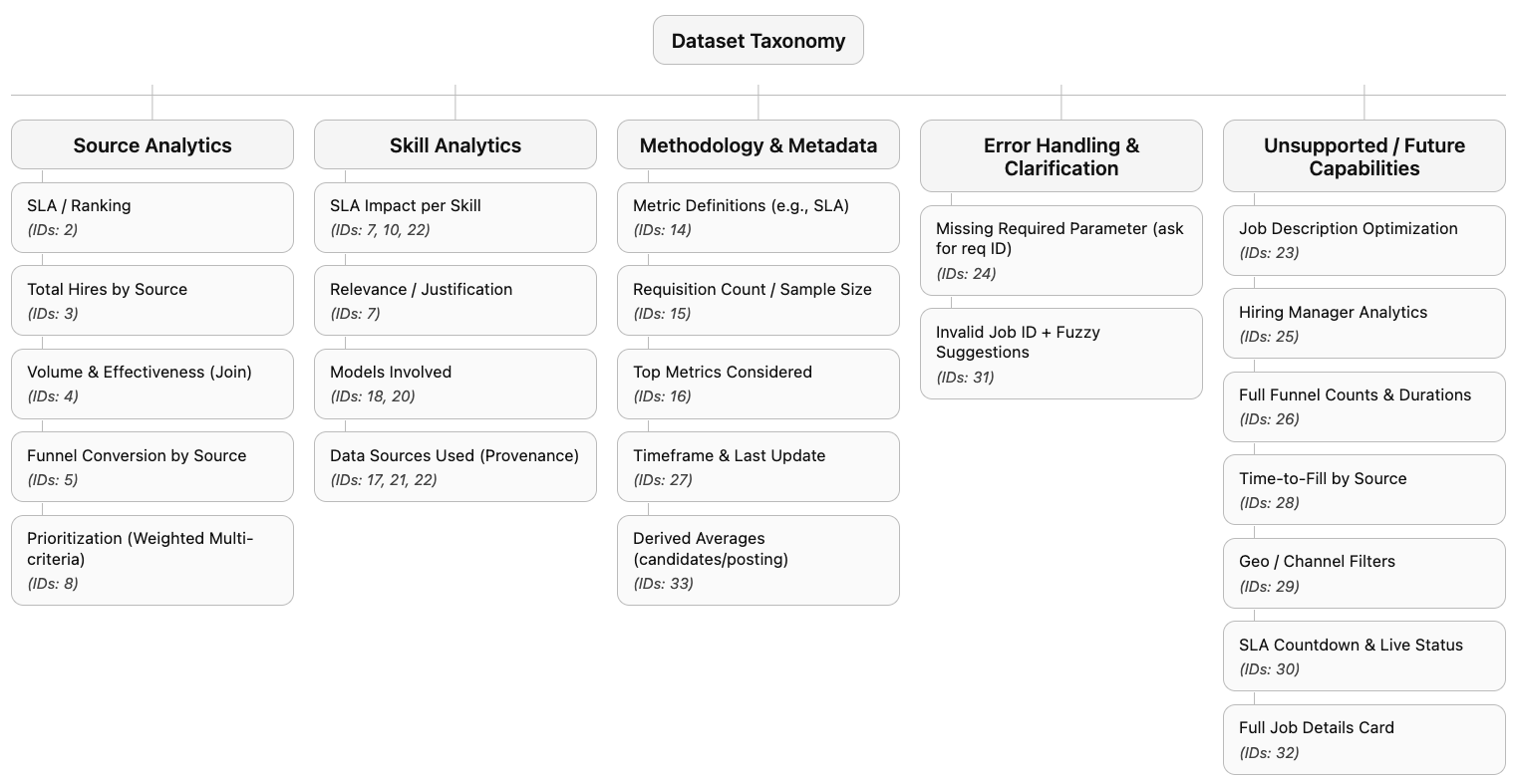}
    \caption{Dataset taxonomy visualization. Each branch corresponds to a query category 
(e.g., Source Analytics, Skill Analytics), and leaf nodes list the dataset entry IDs 
that belong to that category. The visualization highlights the breadth of query types 
covered by the benchmark, ranging from source- and skill-focused analytics to methodology 
and error-handling cases, as well as unsupported or future capabilities.}
    \label{fig:dataset-taxonomy}
\end{figure*}

Figure~\ref{fig:dataset-taxonomy} presents a taxonomy of the dataset catalog, providing a structured overview of the categories represented in our benchmark. The root node ("Dataset Taxonomy") splits into five major categories:

\textbf{Source Analytics:} Queries that focus on requisition sources, including SLA-based ranking, total hires by source, candidate volumes, funnel conversion performance, and weighted prioritization (IDs 2–5, 8).

\textbf{Skill Analytics:} Queries that test reasoning about the impact of skills on SLA performance, fill rates, and relevance judgments, as well as provenance through models and data sources used in calculations (IDs 7, 10, 17, 18, 20, 21, 22).

\textbf{Methodology \& Metadata:} Queries targeting definitional and contextual knowledge, such as SLA definitions, sample sizes, evaluation metrics, timeframe coverage, and averages over requisitions (IDs 14–16, 27, 33).

\textbf{Error Handling \& Clarification:} Queries that probe conversational robustness, such as missing required parameters and invalid job IDs (IDs 24, 31).

\textbf{Unsupported / Future Capabilities:} Queries intentionally designed to highlight system limitations and possible extensions, including job description optimization, hiring manager analytics, full funnel counts, per-source time-to-fill, geographic filters, SLA countdowns, and full job-card details (IDs 23, 25–26, 28–30, 32).

The visualization complements the full catalog table by showing at a glance how queries are distributed across categories. It highlights the dataset's design principle: to balance supported tasks (sources, skills, metadata) with robustness testing (error handling) and aspirational use cases (unsupported/future capabilities).

\subsection{Complete Task Catalog}

Table~\ref{tab:bpota_catalog} lists the full set of 26 BPO-TA tasks, including endpoint usage, glue code requirements, and number of calls. This catalog is intended to ensure reproducibility and provide a reference for ablations.

\begin{table*}[h]
\centering
\scriptsize
\begin{tabular}{p{0.5cm} p{5cm} p{3.5cm} p{3.5cm} p{1cm}}
\toprule
\textbf{ID} & \textbf{Task (Simplified Query)} & \textbf{Endpoints Used} & \textbf{Glue Code / Reasoning} & \textbf{Calls} \\
\midrule
1 & Define SLA metric for requisition 05958BR & \texttt{get\_definitions} & None & 1 \\
2 & Lowest-performing source by SLA for 05958BR & \texttt{recommendation\_summary} & Rank SLA values & 1 \\
3 & Which source had most candidates and how effective? & \texttt{candidate\_volume}, \texttt{recommendation\_summary} & Join on source name & 2 \\
4 & Does Python impact SLA? & \texttt{skill\_impact}, \texttt{recommendation\_summary} & Aggregate by skill & 2 \\
5 & Compare SLA impact of multiple skills (looped) & \texttt{skill\_impact} & Iterative calls per skill, filter negative values & 5+ \\
6 & Which skill most improves SLA? & \texttt{skill\_impact} & Max delta computation & 3+ \\
7 & Funnel conversion for 05958BR & \texttt{funnel\_conversion} & Extract ratios & 1 \\
8 & Candidate drop-off by stage for 05958BR & \texttt{funnel\_conversion} & Compute differences per stage & 1 \\
9 & Top source by time-to-fill (05958BR) & \texttt{recommendation\_summary} & Sort by TTF metric & 1 \\
10 & Cross-source SLA comparison (3 sources) & \texttt{recommendation\_summary} & Filter + compare SLA side-by-side & 1 \\
11 & Explain provenance of Python SLA impact & \texttt{skill\_impact}, \texttt{get\_definitions} & Surface dataset + model metadata & 2 \\
12 & Show data sources used for candidate counts & \texttt{candidate\_volume}, \texttt{get\_definitions} & Join with provenance metadata & 2 \\
13 & Fill rate for source "LinkedIn" on 05958BR & \texttt{recommendation\_summary} & Lookup specific source & 1 \\
14 & Compare conversion rate across 3 requisitions & \texttt{funnel\_conversion} & Aggregate across requisitions & 3 \\
15 & Most stable source across requisitions & \texttt{recommendation\_summary} & Variance analysis across SLA values & 5+ \\
16 & Skill that worsens SLA most & \texttt{skill\_impact} & Min delta computation & 3+ \\
17 & List all requisitions with SLA < 70\% & \texttt{recommendation\_summary} & Filter on threshold & 1 \\
18 & Average SLA by region (unsupported) & None & Must decline gracefully & 0 \\
19 & Hiring manager responsiveness for 05959BR (unsupported) & None & Must decline gracefully & 0 \\
20 & Impact of Wireshark on SLA & \texttt{skill\_impact} & Single lookup & 1 \\
21 & Rank skills by SLA impact for 05958BR & \texttt{skill\_impact} & Sort descending & 5+ \\
22 & Explain SLA computation formula & \texttt{get\_definitions} & Return narrative explanation & 1 \\
23 & Detect missing data for requisition 05959BR & \texttt{recommendation\_summary} & Error handling / graceful failover & 1 \\
24 & Compare top 5 sources for SLA & \texttt{recommendation\_summary} & Sort + slice top-N & 1 \\
25 & Which skill reduces time-to-fill? & \texttt{skill\_impact} & Compare deltas & 3 \\
26 & Unsupported cross-app question (mixing HR + sourcing) & None & Must decline gracefully & 0 \\
\bottomrule
\end{tabular}
\caption{BPO-TA catalog: 26 tasks with associated endpoints, glue code, and number of calls. IDs marked "unsupported" test graceful degradation.}
\label{tab:bpota_catalog}
\end{table*}

\end{document}